\documentclass[conference]{IEEEtran}
\IEEEoverridecommandlockouts
\usepackage{cite}
\usepackage{amsmath,amssymb,amsfonts}
\usepackage{adjustbox, booktabs, multirow, multicol, amssymb, url}
\usepackage{algorithmic}
\usepackage{flushend}
\usepackage{graphicx}
\usepackage{textcomp}
\usepackage{xcolor}
\usepackage{multirow}
\usepackage[flushleft]{threeparttable}
\def\BibTeX{{\rm B\kern-.05em{\sc i\kern-.025em b}\kern-.08em
    T\kern-.1667em\lower.7ex\hbox{E}\kern-.125emX}}

\usepackage{eso-pic}

 \IEEEoverridecommandlockouts
 \IEEEpubid{\makebox[\columnwidth]{
979-8-3503-6486-6/24/\$31.00  
\copyright2024
IEEE \hfill} \hspace{\columnsep}\makebox[\columnwidth]{ }}

\newcommand{\methodname}{RejuvenateFormer}

\begin{document}

\title{An Enhanced Text Compression Approach Using Transformer-based Language Models

}

\author{Chowdhury Mofizur Rahman$^1$ , Mahbub E Sobhani$^2$, Anika Tasnim Rodela$^2$ and Swakkhar Shatabda$^3$\\
$^1$Department of Computer Science and Engineering, State University of Bangladesh \\
$^2$Department of Computer Science and Engineering, United International University\\
$^3$Department of Computer Science and Engineering, BRAC University\\

Email: cmr.cse@sub.edu.bd, msobhani2410011@mscse.uiu.ac.bd, \\ anika@cse.uiu.ac.bd, swakkhar.shatabda@bracu.ac.bd

}

\maketitle

\begin{abstract}
Text compression shrinks textual data while keeping crucial information, eradicating constraints on storage, bandwidth, and computational efficacy. The integration of lossless compression techniques with transformer-based text decompression has received negligible attention, despite the increasing volume of English text data in communication. The primary barrier in advancing text compression and restoration involves optimizing transformer-based approaches with efficient pre-processing and integrating lossless compression algorithms, that remained unresolved in the prior attempts. Here, we propose a transformer-based method named {\methodname} for text decompression, addressing prior issues by harnessing a new pre-processing technique and a lossless compression method. Our meticulous pre-processing technique incorporating the Lempel-Ziv-Welch algorithm achieves compression ratios of 12.57, 13.38, and 11.42 on the BookCorpus, EN-DE, and EN-FR corpora, thus showing state-of-the-art compression ratios compared to other deep learning and traditional approaches. Furthermore,  the {\methodname} achieves a BLEU score of 27.31, 25.78, and 50.45 on the EN-DE, EN-FR, and BookCorpus corpora, showcasing its comprehensive efficacy. In contrast, the pre-trained T5-Small exhibits better performance over prior state-of-the-art models.
\end{abstract}


\section{Introduction}
Text compression, a fundamental aspect of data compression, refers to the process of reducing the size of textual data yet preserving its important information content. In simple terms, it indicates encoding text to represent it substantially smaller in size, thereby optimizing storage space, transmission bandwidth, and computational resources. The increasing amount of textual data generated and shared in a broad spectrum of domains, such as communication networks, digital libraries, web content, and document management systems, makes text compression indispensable. Worldwide, 18.7 billion texts are transmitted each day, not including messages exchanged between apps \cite{textVermont}. By 2022, 95\% of businesses shifted from paper to digital documentation, increasing operational efficiency and minimizing security risks. Because of the scalability and security, cloud-based document management systems are utilized by 98\% of tech companies \cite{DocumentStat}. Due to the exponential growth of digital content, robust solutions for compressing text without losing its integrity or readability are important. Text compression techniques address this issue by ensuring efficient storage, retrieval, and transmission of textual data, thereby enhancing information systems' scalability, performance, and cost-effectiveness. Additionally, text compression has become vital for enabling efficient data transfer and interoperability across heterogeneous systems and platforms.

A range of approaches can be found in text compression techniques to reduce the volume of text data while maintaining its important information content. Lossless compression techniques, such as the Lempel-Ziv-Welch (LZW) \cite{nelson1989lzw} algorithm and Huffman coding \cite{knuth1985dynamic}, leverage patterns and redundancies in the text to produce compact representations while achieving compression without losing any data. Conversely, lossy compression techniques, such as statistical and dictionary-based approaches, trade off some quality in exchange for greater compression ratios. Recently, deep learning techniques have become efficient means for attaining advanced compression efficiency. These models may automatically find effective representations and encoding schemes by training on large corpora of text data. This enables greater compression performance over conventional techniques. The impact of transformer-based models—like BERT \cite{li2023crossword}, LLaMA \cite{valmeekam2023llmzip}, and ALBERT \cite{li2021text}—has been emphasized in earlier research. These models have proven to be successful in preserving contextual information for both high and low-resource languages during text decompression. Architectural changes exhibiting good performance, such as the BBFNMT \cite{li2020explicit}, are noteworthy. Not to mention, following the trend of refining domain-specific pre-trained models and integrating the LoRA \cite{hu2021lora} technique has also produced excellent outcomes \cite{ge2023context}. Cross-lingual augmentation techniques improve transformer models' performance for languages based on various resource availability—a topic that was not addressed in the research of Mao et al. \cite{mao2022trace}. The study of Huang et al. \cite{huang2023approximating} utilized the operational process of arithmetic coding and combined it with GPT for lossless text compression which was more of a complex approach with additional computational latency. For a comprehensive assessment of compression ratios, at least three general-purpose compressors must be considered. It is also crucial to keep in mind that iterative assessment of meaning preservation may take time, and non-autoregressive decoding may not restore the original text flawlessly \cite{ge2022lossless}.

We addressed several constraints associated with text compression, especially concerning the lack of a futuristic approach that integrates lossless compressors and transformer-based methods. In this study, we propose a simple yet effective pre-processing technique to improve the compression ratio and restore the original text by leveraging a transformer-based method called {\methodname}, wherein we optimize complexity and computational efficiency through a tailored architecture consisting of six encoder and decoder layers. The contribution of our study is summarized below:

\begin{itemize}
    \item A simple strategic pre-processing technique tailored for the English language has been introduced, aimed to improve text compression ratios.
    \item After thorough pre-processing, various general-purpose lossless compression methods are utilized across multiple corpora to evaluate the effectiveness of our technique in terms of compression ratio. Notably, our method demonstrated state-of-the-art results, with the Lempel-Ziv-Welch (LZW) algorithm showcasing exceptional performance.
    \item  The pre-trained T5-Small showcases adequate performance over {\methodname} and BBFNMT; underscoring our approach of strategic pre-processing incorporating transformer-based encoder-decoder models has the potential to attain benchmark results; this paved the way for large-scale text restoration experiments.
    \item An exploration into the impact of the size of the training corpus on the efficacy of {\methodname} in restoring missing characters within English text is conducted, shedding light on its potential contributions in the realm of decompression.
\end{itemize}

The following sections of the paper are organized as follows: an exhaustive review of the most relevant prior research on text compression. Next, we explain our text pre-processing technique. The subsequent section elucidates our proposed architecture. Afterward, we present a detailed account of the experimental analysis, and, finally, we offer concluding remarks and discuss future research avenues.

\section{Related Work}
In recent years, we have experienced an upsurge of digital documents in a broad spectrum of domains, including government, business, healthcare, and academia. Such rapid changes have led to an extreme need for reliable compression algorithms that can reduce data sizes without risking the content's quality and integrity.  This literature review presents an in-depth evaluation of document compression techniques for lossless and lossy methods, ranging from traditional to trendy techniques while highlighting the methodologies, guiding principles, benefits, and limitations.

Transformer-based techniques, utilizing GPT, LSTM, and a single-layer transformer, have been widely utilized for text compression. Valmeekam et al. \cite{valmeekam2023llmzip} propose a novel study that leverages large language models (LLMs) such as LLaMA-7B\cite{touvron2023llama} to present a novel approach for lossless text compression. Using a window of prior tokens, the method utilizes the LLM to determine the next token (based on probability ranking) in a text sequence. The rank or probability of the actual token is then encoded using a lossless compression scheme, such as zlib, token-by-token, or arithmetic coding. Subsequently, harnessing the decoder-only transformer Schmidt et al. \cite{schmidt2024tokenization} developed PathPiece, a new tokenizer that splits a document's content into the minimum number of tokens necessary for specific vocabulary, to test the hypothesis that fewer tokens improve downstream performance. In this study, the authors trained 64 language models utilizing distinct tokenization and parameter sizes ranging from 350M to 2.4B.

One of the LSTM-based techniques that Goyal et al. \cite{goyal2018deepzip} proposed is a technique that uses recurrent neural networks, for prediction and arithmetic coder algorithms for lossless compression that achieves around ideal compression for synthetic datasets, outperforming Gzip on real datasets. In contrast, the LSTM-based model can perform significantly well on classification tasks \cite{rodela2023bangla}. 

Among lossy text compression, Mingxiao et al. \cite{li2023crossword} presents an innovative text compression strategy that offers superior compression performance and semantic stability by masking fewer significant words and restoring them through a transformer-based model. The masked text is converted into a bit string using LZW encoding, and then sentence-BERT evaluates semantic importance. Masked words can be restored using a transformer-based demasking module, lowering the cross-entropy between ground truth and soft decisions. Whereas, Zuchao et al. \cite{li2021text} proposed an approach whose goal was to extract the core information of the input text by improving the text encoding of transformer models with text compression using two different approaches. They are Explicit Text Compression (ETC) and Implicit Text Compression (ITC). The basic idea of ITC is to produce compressed text features without generating an explicit text sequence through the use of a non-autoregressive decoder.

Amid BERT-based methods, Kale et al. \cite{kale2024texshape} used neural networks for mutual information estimation and information-theoretic compression in addition to a benchmark language model for initial text representation. In terms of the predictive accuracy of downstream models trained on the compressed data, their studies show notable improvements in keeping maximal targeted information and minimal sensitive information over detrimental compression ratios. Furthermore, Zhang et al. \cite{zhang2024dc} offer the DC-Graph model to overcome the issues with reconstructing and augmenting information in lengthy texts incorporating different optimized modules with pre-trained BERT. Their proposed architecture outperforms Recurrent Chunking Mechanisms and BERT.

Anisimov et al. \cite{anisimov2024natural} created a basic monotonic mapping from the set of non-negative integers to the codeword set, which is utilized to construct a fast byte-aligned decoding algorithm showcasing a better compression ratio when compared to known codes of the same type. 

Khan et al.\cite{khan2024graph} proposed a method that combines Named Entity Recognition (NER) and Graph Neural Networks (GNNs) for text summarization, focusing on important document structures and relevant entities. For summarizing large text, this integration ensures both relevance and efficiency.

Aslanyurek et al.\cite{aslanyurek2023new}  introduced WSDC (Word-based Static Dictionary Compression) and used iterative clustering to build static dictionaries for high compression ratios in short texts. DSWF (Dictionary Selection by Word Frequency) is proposed to select the most suitable dictionary for compressing the source text. WSDC shows superior performance than other methods for short texts under 200 and 1000 bytes(except Zstd).

\begin{figure*}[t] 
  \centering
  \includegraphics[width=0.8\textwidth]{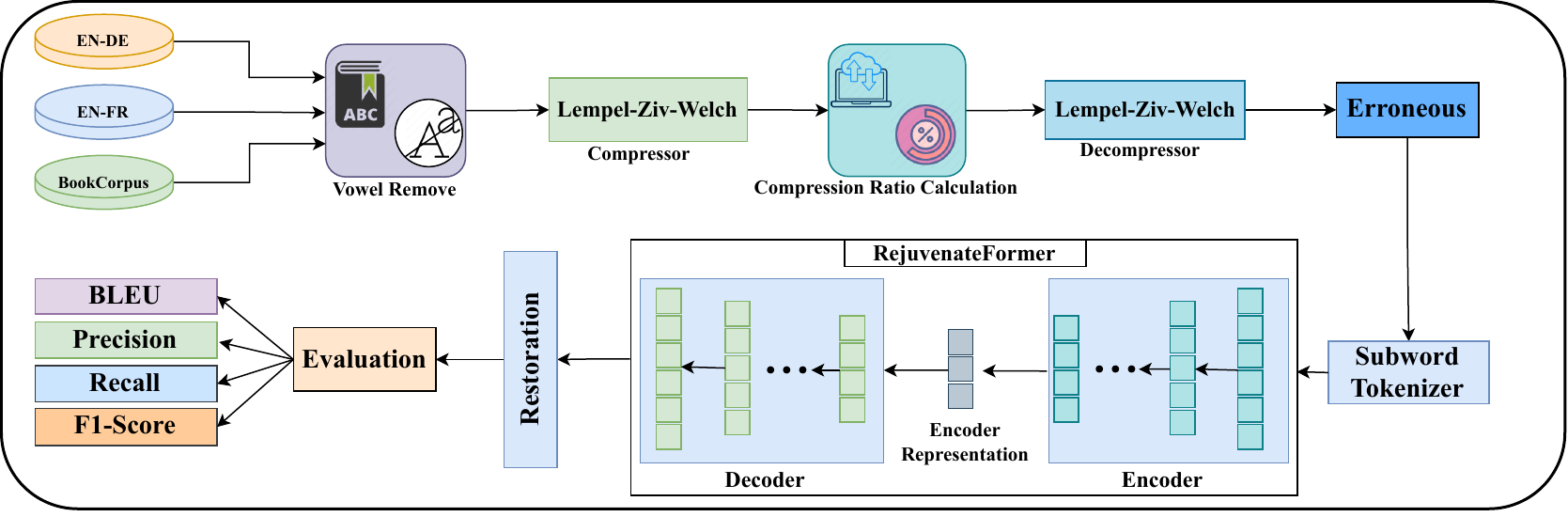} 
  \caption{\textbf{(Top)} Each corpus undergoes vowel removal, and the compression ratio is calculated using the compressed representation. The text is then reverted to its earliest form without vowels. \textbf{(Bottom)} After tokenization, the {{\methodname}} is trained on each corpus, to proficiently generate expected outcomes.}
  \label{fig:your-image-label}
\end{figure*}

\section{Dataset}
\subsection{Data Sourcing}
The source of our data is from two publicly available standard large-scale corpora: WMT14 and BookCorpus. Within the WMT14, EN-DE, and EN-FR language pairs datasets comprise 1.5 million and 1.6 million sentence pairs respectively and the BookCorpus dataset contains 7.8 million English sentences.

\subsection{Data Preprocessing}
Beneath the WMT14 dataset, we consider English-to-German and English-to-French language pairs. We only extracted the column containing English text from both the dataset. In the BookCorpus dataset, it was not necessary to do this procedure as it has a single column containing English text. Then, we identified 5 vowel letters, in both capital and small letters, that appear in the English text. These vowel letters are denoted as $VC = \{VC_1, VC_2, ..., VC_{10}\}$. Next, we take every sentence indicated as $S = \{S_1, S_2, ..., S_{N}\}$, where $N$ indicates the number of characters in the sentence. We traverse through each of the characters $S_i \in S$ and remove any character that is present in the character set $VC$. By following the above process, the dataset has been constructed in source-target pairs. In this source-target pair, the source sentence has no vowel letters, while the target sentences are corrected versions with missing vowel letters.

\section{Proposed Method}

%

\subsection{Problem Formulation \& Overview}
Consider input sentences $S = \{S_1, S_2, \ldots, S_{N}\}$, where $N$ is the number of character. The vowel remover ($VR(\cdot)$) takes the input sentence and removes the vowels. The Lempel–Ziv–Welch compressor ($LZWC(\cdot)$) takes the sentence and the compressed representation is stored in a hard disk, whereupon calculations regarding length are computed. The Lempel–Ziv–Welch decompressor ($LZWD(\cdot)$) decompresses the compressed representation. Subsequently, consider pair of tokens, $SRC_{I} = \{src_1, src_2, \ldots, src_{n}\}$ and $TGT_{I} = \{tgt_1, tgt_2, \ldots, tgt_{k}\}$, where  $SRC_{I}$ represents the input sequence that has no vowel, and $TGT_{I}$ represents the target sequence with vowel restored. Erroneous input sequence $SRC_{I}$ is tokenized using a subword tokenizer ($T(\cdot)$) and then fed into the encoder ($E(\cdot)$) and generates context vectors of the sentence, denoted as $CV = \{CV_1, CV_2, \ldots, CV_{512}\}$. The decoder ($D(\cdot)$) utilized the context vector $CV$, alongside the previously decoder generated tokens to generate the correct sentence autoregressively. The pre-processing can integrate seamlessly with any transformers-based encoder-decoder models. The entire procedure can be mathematically abbreviated as follows:
\begin{equation}
    SRC_{I}=LZWD(LZWC(VR(S)))
    \label{eq:src}
\end{equation}
\begin{equation}
    \hat{TGT} = D((E(T([SRC_I]), W^E), D_{out}^{t-1}), W^D)
    \label{eq:method}
\end{equation}

\subsection{{\methodname}}
In this section, we discuss the details of {\methodname}.

\subsubsection{\textbf{Lempel-Ziv-Welch}}
Lempel-Ziv-Welch (LZW) compression forms a dictionary of symbols, replaces repeated substrings with dictionary indices, and updates the dictionary during compression. During decompression, it initializes, reads, retrieves, and updates the dictionary to reconstruct the original data.

\subsubsection{\textbf{Vowel Removal and Compression}}
In our text processing pipeline, the function ($VR(\cdot)$) removes vowels from input sentences, producing a text devoid of vowels. Subsequently, the Lempel-Ziv-Welch compressor ($LZWC(\cdot)$) generates a compressed representation of the text, which is then used to calculate the length optimization ratio. Lastly, the compressed representation is decompressed using the ($LZWD(\cdot)$), restoring the original text without vowels, which is then fed into our transformer model to recover the vowels.

\subsubsection{\textbf{Encoder}}
We transform each sentence into a sequence of tokens $\mathbf{X} = \{x_1, x_2, \ldots, x_{n}\}$ and assign discrete values to each subword. Here $n$ is the sequence length. We confirmed the same input dimension to every input sequence by incorporating padding to $\mathbf{X}_i$. Subsequently, each token $x_{i}$ went through an embedding layer $\mathbf{E}$ to transform the discrete values to continuous vectors which represent trainable matrix $\mathbf{E}{x_i} = \mathbf{Embedding}{(x_i)}$. Through backpropagation during training, these metrics are fine-tuned to minimize the loss. To preserve the positional order of the tokens in a sentence we incorporated positional encoding $\mathbf{PE}$ for each token $x_{i}$, denoted as $\mathbf{PE}{x_i}$ with the embedding. This composite embedding, represented as $\mathbf{Z}{x_i} = \mathbf{E}{x_i} + \mathbf{PE}{x_i}$ is then fed into the stack of $\mathbf{K}$ identical encoder layers which are built upon two main components: a position-wise-feed-forward network and a multi-head-self-attention layer. Furthermore, the self-attention mechanism calculates attention score for each token by looking into all other tokens in a sequence $\mathbf{X}$ using three learnable vectors, query ($\mathbf{Q}$), key ($\mathbf{K}$), and value ($\mathbf{V}$) for capturing the contextual dependencies in a sequence of tokens. The mathematical expression of self-attention is defined as follows \cite{vaswani2017attention}: 
\begin{equation}
    \resizebox{.80\hsize}{!}{$\mathbf{Attention}(\mathbf{Q}, \mathbf{K}, \mathbf{V}) = \mathbf{softmax}(\frac{\mathbf{QK}^T}{\sqrt{\mathbf{d_k}}}) \mathbf V$}
    \label{eq:self_attention}
\end{equation}

The position-wise-feed-forward mechanism introduces non-linearity to the encoder through two linear transformations by incorporating ReLU, which is a non-linear activation function for enhancing the obtained representation. The output of each encoder layer is propagated sequentially to $\mathbf{K}$ identical layers which output rich contextualized representation of input sequence $\mathbf{X}$ covering local and global dependencies.

\subsubsection{\textbf{Decoder}}
The target sequence, $\mathbf{Y} = \{y_1, y_2, \ldots, y_{m}\}$, where $m$ is the sequence length, undergoes through an embedding layer $\mathbf{E}{y_i} = \mathbf{{Embed}(y_i)}$ for converting tokens with discrete values into continuous vectors. For preserving positional information of tokens in a sequence, positional encoding $\mathbf{PE}{y_i}$ in blended with embedding $\mathbf{E}{y_i}$ and the resulting embedding is $\mathbf{Z}{y_i} = \mathbf{E}{y_i} + \mathbf{PE}{y_i}$. This combined embedding is passed through $\mathbf{L}$ identical decoder layers. The decoder layers are composed of two main components: a position-wise-feed-forward layer and a masked multi-head-self-attention layer. Masking is incorporated in the decoder so that it can not generate the current token by attending the future tokens. Following this distinction, the computational process of masked-multi-head-self-attention follows the same equation \eqref{eq:self_attention} as the encoder's muti-head-self-attention. Furthermore, the position-wise-feed-forward network enhances the learned representation by following the same procedure as the encoder's position-wise-feed-forward network. The acquired learned representation undergoes through $\mathbf{L}$ identical decoder layers and produces the fine-graded target sequence.

\renewcommand{\arraystretch}{1.4}
\begin{table*}[t]
\caption{The comparison of the quantitative performance of different existing methods across different datasets.}
    \centering
    \begin{adjustbox}{width=0.85\textwidth}
    \begin{tabular}{|l|c||cccc||cccc||cccc||}
        \hline
        \multirow{2}{*}{\textbf{Method}} & \multirow{2}{*}{\textbf{\#Params.}} & \multicolumn{4}{c||}{WMT14 EN-DE} & \multicolumn{4}{c||}{WMT14 EN-FR} & \multicolumn{4}{c||}{BookCorpus} \\
        \cline{3-14}
         {} & {} & \textbf{BLEU} & \textbf{PR} & \textbf{RE} & \textbf{F1} & \textbf{BLEU} & \textbf{PR} & \textbf{RE} & \textbf{F1} & \textbf{BLEU} & \textbf{PR} & \textbf{RE} & \textbf{F1}\\
         \cline{1-14}

         BBFNMT \cite{li2020explicit} & \textbf{230.3M} & 29.37 & $-$ & $-$ & $-$ & 42.52 & $-$ & $-$ & $-$ & $-$ & $-$ & $-$ & $-$  \\
         T5-Small & {60.51M} & 41.44 & 0.93169 & 0.91312 & 0.92211 & 34.57 & 0.9224 & 0.9039 & 0.9129 & 63.61 & 0.9703 & 0.9539 & 0.9618  \\
         \textbf{{\methodname}} & 63.23M & \textbf{27.31} & \textbf{0.8717} & \textbf{0.9139} & \textbf{0.8921} & \textbf{25.78} & \textbf{0.8691} & \textbf{0.9131} & \textbf{0.8903} & \textbf{50.45} & \textbf{0.9384} & \textbf{0.9527} & \textbf{0.9454}  \\
        \hline
    \end{tabular}
    \end{adjustbox}
    \label{tab:quantitative_res}
\end{table*}

\subsubsection{\textbf{Hyperparameters}}
The hidden dimension is kept as 512 through all the encoder and decoder layers for maintaining consistency. To maintain the model's depth and capacity the number of neurons is kept at 2048 for the feed-forward layer and a 0.1 dropout ratio is applied to prevent overfitting. To maintain the efficient computation and non-linearity over the network we have incorporated ReLU. The model went through 50 epochs with $5 \times 10^{-5}$ learning rate incorporating AdamW optimizer. We incorporated the categorical cross-entropy loss function for the optimization process, which leads the model towards desired translations.

\section{Experimental Analysis}
\subsection{Datasets}
Because of limited computational resources, we selected 100K sentence pairs from each of the three datasets.
\begin{itemize}
    \item \textbf{WMT14 \cite{bojar2014findings}}
    The WMT14 dataset was used as shared tasks of the Ninth Workshop on Statistical Machine Translation. Within five language pairs, we used English-to-French (EN-FR) and English-to-German (EN-DE) datasets, which are both large-scale standard corpus.
    \begin{itemize}
        \item \textbf{English-French.}
        It is composed of 1.6M source-target pairs after our rigorous pre-processing. The corpus was divided into training, validation (newstest2013), and test (newstest2014) sets. As a result, the training, validation, and test sets comprise 100000 (100K), 3000 (3K), and 3003 (3k) source-target pairs respectively.
        \item \textbf{English-German.}
        The dataset contains a total of 1.5M source-target pairs after our thorough pre-processing. We separated the corpus into training (100K), validation, and test sets, keeping newstest2013 as the validation set and newstest2014 as the data in the test set. Accordingly, the resulting training, validation, and test set comprise 100000 (100K), 3000 (3K), and 3003 (3k) source-target pairs respectively.
    \end{itemize}
    \item \textbf{BookCorpus \cite{zhu2015aligning}}
    An extensive set of free novel books, mostly authored by unpublished authors are included in the BookCorpus dataset. Following a meticulous text pre-processing phase, we found a total of 7.8M source-target pairs within the corpus. Consequently, we partitioned the corpus into 100000 (100K) pairs in the training and 5000 pairs in the test set to create distinct training and test sets.
\end{itemize}

\subsection{Baselines}
\begin{itemize}
    \item \textbf{T5-Small \cite{raffel2020exploring}} Text-to-Text Transformer (T5) is a language model developed by Google. T5-Small is the smaller variant of T5 consisting of ($\approx$70M) parameters where the base version is of (220M) parameters. The T5-Small was created aiming to maintain good performance with a smaller number of parameters.
\end{itemize}

\subsection{Performance Evaluation}
We evaluated the performance of our approach with compression ratio (\textbf{Original Length/Compressed Length}). We also evaluated the performance of our model in restoring vowels by calculating BERTScore \eqref{eq:f1} and BLEU \eqref{eq:sacreBLEU} scores.

\begin{itemize}
    
    \item \textbf{BERTScore. \cite{bert-score}}
    The BERTScore calculates the semantic similarity of two pieces of text by calculating the cosine similarity of their embedding tokens. This metric outputs precision, recall, and f1 score, and their equations are as follows:
    \begin{equation}
        Recall_{\mathrm{BERT}}= \frac{1}{N}\times \sum_{1}^{N}\left( \frac{1}{|x|} \sum_{x_i \in x} \max _{\hat{x}_j \in \hat{x}} \mathbf{x}_i^{\top} \hat{\mathbf{x}}_j \right)
        \label{eq:recall}
    \end{equation}
    \begin{equation}
        Precision_{\mathrm{BERT}}= \frac{1}{N}\times \sum_{1}^{N}\left( \frac{1}{|\hat{x}|} \sum_{\hat{x}_j \in \hat{x}} \max _{x_i \in x} \mathbf{x}_i^{\top} \hat{\mathbf{x}}_j \right)
        \label{eq:precision}
    \end{equation}
    \begin{equation}
        F1_{BERT} = \frac{1}{N}\times \sum_{1}^{N} (2\times \frac{P_{BERT} \times R_{BERT}}{P_{BERT}+R_{BERT}})\
        \label{eq:f1}
    \end{equation}
    where $\mathbf{x}_i^{\top} \hat{\mathbf{x}}_j$ is the cosine similarity between two pieces of text and N is the total number of sentence pairs. 

    \item \textbf{BLEU. \cite{post-2018-call}}
    The BLEU metric estimates the quality of candidate text by assigning precision scores to n-grams and comparing them with one or more reference texts. Scores range from 0 and 100, where a higher score denotes better results. The mathematical formula for BLEU is as follows:
    \begin{equation}
        \text{BLEU} = \text{BP} \times e^{\sum_{n=1}^{N}(w_{n}\cdot logp_{n})}
        \label{eq:sacreBLEU}
    \end{equation}

    Here, The Brevity Penalty (BP) punishes shorter predictions. $N$ is the maximum n-gram length. $w_{n}$ are weights for n-gram precision, and log$p_{n}$ is the logarithm of n-gram precision in the candidate text.
\end{itemize}

\subsection{Experimental Results}


        

\subsubsection{\textbf{Quantitative Results}}
We evaluated the results of our pre-processing technique on all three corpora. We tested the significance of removing vowels in compressing text using various general-purpose lossless compressors such as LZMA, LZW, GZIP \cite{deutsch1996gzip}, ZLIB \cite{deutsch1996zlib}, and Arithmetic Coding (AC) \cite{witten1987arithmetic}. Our rigorous experiment found that LZW compresses text 13.38$\times$, 12.57$\times$, and 11.42$\times$ on EN-DE, BookCorpus, and EN-FR corpus respectively, and establishes itself as the state-of-the-art in compression ratio by outperforming GPT-based approach \cite{huang2023approximating}, TRACE \cite{mao2022trace}, and other compressors. Table \ref{tab:com_ratio} demonstrates that removing vowels significantly improves the compression ratio and lowers the length incorporating 
compressed representation. In contrast, we presented the quantitative performance of different Transformer-based methods in Table \ref{tab:quantitative_res}. Our proposed model, {\methodname}, shows favorable performance across all three corpora in small-scale experiments being 3.6$\times$ small in terms of parameter count. In contrast, the pre-trained T5-Small performed significantly well over BBFNMT
\renewcommand{\arraystretch}{1.4}
\begin{table}[h!]
    \caption{The juxtaposition of the compression ratio of different existing methods across various corpora.}
    \centering
    \begin{adjustbox}{width=0.47\textwidth}
    \begin{tabular}{lcccccccc}
        \hline
        \multirow{2}{*}{\textbf{Dataset}} & \multirow{2}{*}{\textbf{Corpus}} & \multicolumn{5}{c}{{Vowel Remove (Ours)}} & \multirow{2}{*}{\textbf{GPT}} & \multirow{2}{*}{\textbf{TRACE}}\\
        \cline{3-7}
        & {\textbf{Size}} & \textbf{LZW} & \textbf{LZMA} & \textbf{GZIP} & \textbf{GLIB} & \textbf{AC} & {} & {} \\
        \hline
        \centering {BookCorpus} & \centering 6.7M & \textbf{12.57} & 4.583 & 3.603 & 3.575 & 2.907 & 10.55 & 4.49 \\
        \centering {EN-DE} & \centering 1.5M & \textbf{13.38} & 5.257 & 3.959 & 3.927 & 2.947 & $-$ & $-$ \\
        \centering {EN-FR} & \centering 1.6M & \textbf{11.42} & 4.656 & 3.643 & 3.626 & 2.953 & $-$ & $-$ \\
        \hline \\
\end{tabular}
    \end{adjustbox}
    \label{tab:com_ratio}
\end{table}
and {\methodname} across all evaluation metrics and corpora, except for the EN-FR dataset with only 40K sentence pairs and 30 epochs. This highlights the effectiveness of larger contextual models when integrated with our proposed text pre-processing strategy. However, we did not consider the scores for each sentence separately; instead, we calculated the overall BLEU and BERTScore by evaluating all the sentences in a corpus.

\subsubsection{\textbf{Qualitative Results}}
We evaluated the qualitative performance of T5-Small and {\methodname}. Our proposed model showed superior performance in both training and inference times compared to T5-Small in a variety of sentence conditions. We found that while T5-Small was accurate in restoring short sentences with a low number of missing vowels, it struggled with longer sentences that had more missing vowels. However, due to its training on a larger corpus, T5-Small outperformed {\methodname} in terms of the overall score, benefiting from a wider linguistic context. In contrast, our {\methodname} model demonstrated its competitive performance by accurately restoring sentences with a higher number of missing vowels in different sentence structures.

\subsubsection{\textbf{Ablation Study}}
Table \ref{tab:ablation_study}
\renewcommand{\arraystretch}{1.4}
\begin{table}[h!]
\caption{The influence of corpus size (EN-DE) on the performance of our proposed method.}
\centering
\begin{adjustbox}{width=0.47\textwidth}
\begin{tabular}{lccccc}
\hline
\multirow{2}{*}{\textbf{Method}} & \multirow{2}{*}{\textbf{Corpus}} & \multicolumn{4}{c}{{Inference}} \\
\cline{3-6}
& {\textbf{Size}} & \textbf{BLEU} & \textbf{PR} & \textbf{RE} & \textbf{F1} \\
\hline
\centering {{\methodname}} & \centering 30K & 23.07 & 0.8714 & 0.9012 & 0.8859 \\
\centering {{\methodname}} & \centering 50K & 25.12 & 0.8720 & 0.9071 & 0.8890 \\
\centering {{\methodname}} & \centering 100K & 27.31 & 0.8717 & 0.9139 & 0.8921 \\
\hline \\
\end{tabular}
\end{adjustbox}
\label{tab:ablation_study}
\end{table}
shows the impact of corpus size on model performance. In this extensive study, we used three large-scale datasets but the correlation between the size of the corpus and the effectiveness of the model performance has been shown for the WMT14 EN-DE language pair dataset. Due to computational constraints, we limited our experiment to a smaller number of sentence pairs. Surprisingly, the corpus with 100K instances outperformed those with 30K and 50K instances. Conversely, the corpus with 30K instances showed the least significant results, while the 50K instances corpus delivered moderate outcomes. This tendency was seen in all three datasets, signifying that larger corpus sizes lead to enhanced performance \cite{bijoy2023advancing}.

\section{Conclusion}
This study identified the primary obstacle in text compression and proposed a comprehensive architecture. We introduced a new pre-processing technique that leverages the compatibility of the Lempel-Ziv-Welch algorithm to reduce the length of text, resulting in a state-of-the-art compression ratio on EN-DE, EN-FR, and BookCorpus dataset. Concurrently, we validated the efficacy of our proposed technique on these three large-scale corpora, confirming its compatibility with transformers-based encoder-decoder architecture like pre-trained T5-small and establishing it as a promising method for text restoration tasks. Notably, we introduced a transformer-based text restoration method, meticulously designed to address complex linguistic patterns by harnessing attention mechanisms and positional encoding resulting in promising evaluation scores even in small-scale experiments. In our future study, we will incorporate our transformer-based method into large-scale training and experiments, and evaluate the efficiency of knowledge distillation.
\bibliographystyle{ieeetr}
\bibliography{reference}
\vspace{12pt}
\end{document}